\documentclass[10pt,twocolumn]{article}
\usepackage[table]{xcolor}

\usepackage{times}
\usepackage[margin=2cm]{geometry}
\usepackage{amsmath,amssymb,amsthm}
\usepackage{graphicx}
\usepackage{booktabs}
\usepackage{multirow}
\usepackage{xspace}
\usepackage{algorithm}
\usepackage{algorithmic}
\usepackage[numbers,sort&compress]{natbib}
\usepackage{hyperref}
\usepackage{caption}
\usepackage{subcaption}
\usepackage{array}
\usepackage{enumitem}
\usepackage{bm}
\usepackage{float}

\definecolor{tableheader}{RGB}{46,134,171}
\definecolor{tablerowalt}{RGB}{245,248,250}
\definecolor{bestresult}{RGB}{46,134,171}

\title{Cross-Modal Hierarchical Fusion for from Multi-Sensor Ground Observations}

\author{
  Xinze Zhang$^{1}$  \\[6pt]
  $^{1}$University of Southern California, Los Angeles, CA 90007, USA \\[3pt]
  \quad $^{*}$Corresponding author: zhangxinze00@outlook.com
}

\date{}

\begin{document}
\maketitle

% ==============================================================================
% ABSTRACT
% ==============================================================================
\begin{abstract}
Dense volumetric reconstruction of cloud microphysical fields from sparse ground-based instruments remains an open problem, largely because the available measurements are heterogeneous in both modality and spatial coverage. We present \textbf{AtmoFuseNet}, a framework that fuses multi-view sky camera imagery with millimeter-wave cloud radar and ceilometer observations to produce 4D (three spatial dimensions plus time) estimates of cloud state and wind. The method operates in three stages: a cross-modal hierarchical aggregation module that combines image feature pyramids with instrument-derived vertical profiles through layer-wise cross-attention; a conditional variational refinement module that maps the resulting volume to physically consistent microphysical fields under differentiable radar and image forward models; and a correlation-based motion estimator that recovers per-voxel 3D wind vectors from consecutive volumetric reconstructions. On collocated observations from a semi-arid site, AtmoFuseNet reaches 0.026~g~m$^{-3}$ liquid water content MAE and 1.18~m~s$^{-1}$ wind speed MAE, improving over existing retrieval baselines. Ablation experiments isolate the contribution of each module.
\end{abstract}


%%%%%%%%%%%%%%%%%%%%%%%%%%%%%%%%%%%%%%%%%%
\section{Introduction}

The radiative and hydrological role of clouds in Earth's climate system is well established, yet their internal microphysical structure continues to be poorly constrained by observations~\cite{INTENT,HABIT,yu2025physics,li2026retrack}. Shallow cumulus clouds present a particularly acute challenge: they span only a few hundred meters horizontally, evolve within minutes, and occupy spatial scales well below the grid spacing of current satellite products and global reanalysis datasets~\cite{hersbach2020era5,REFINE,li2025exploring,li2025stitchfusion,sarkar2025reasoning,chen2026intent}. Because numerical weather prediction and climate models must parameterize rather than resolve these sub-grid processes, the accuracy of such parameterizations is difficult to assess without spatially dense ground-truth observations~\cite{li2025stitchfusion,li2025maris,yu2026spatiotemporal,li2026habit}.

Ground-based platforms partially alleviate this problem. Multi-camera sky imaging arrays deliver panoramic views at cadences of a few seconds, while zenith-pointing cloud radars and ceilometers supply vertically resolved reflectivity profiles and cloud boundary heights~\cite{li2025exploring2,li2025u3m,yu2026dinov3,yu2025qrs,fu2026airknow}. Yet each of these sensors covers only part of the information needed for a full volumetric reconstruction: cameras capture horizontal structure but lack depth, whereas profiling instruments sample along a single vertical column. Combining these complementary modalities into a spatially coherent 3D picture therefore requires careful cross-modal alignment together with appropriate physical constraints.

\begin{figure*}
\centering
\includegraphics[width=0.99\textwidth]{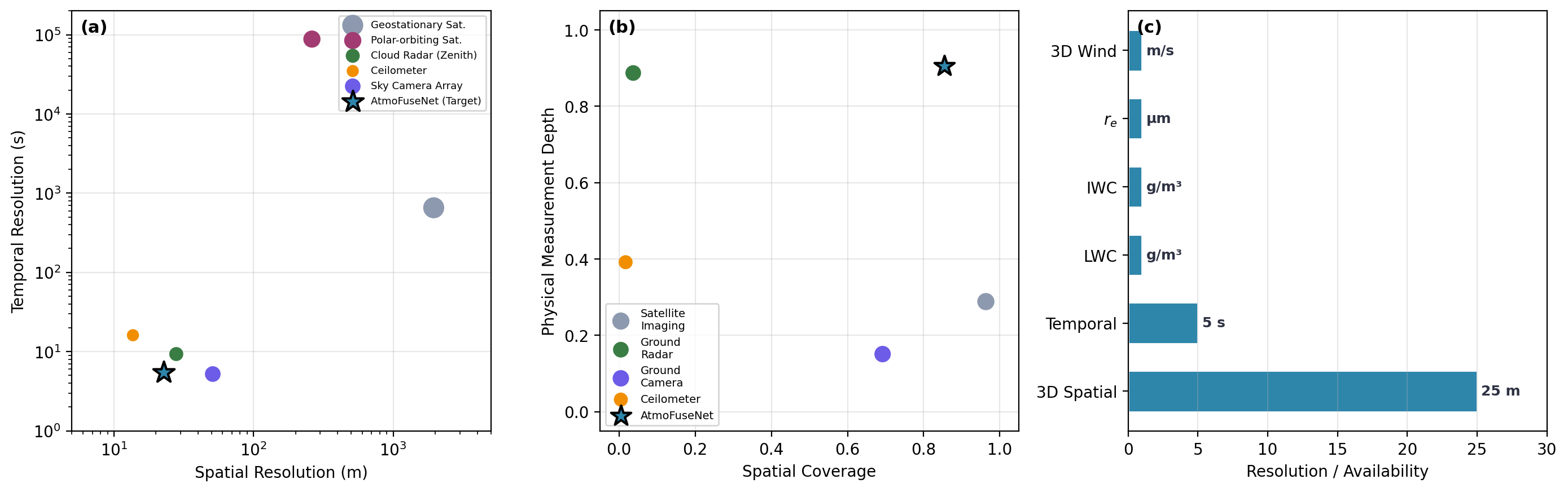}
\caption{Multi-sensor observation gap motivating this work. (a) Spatial versus temporal resolution of individual instruments; the star marks the target achieved by AtmoFuseNet. (b) Trade-off between horizontal coverage and physical measurement depth. (c) Target 4D output channels.}
\label{fig:motivation}
\end{figure*}

Neural scene representations built on differentiable rendering have recently shown that volumetric fields can be recovered from a handful of images~\cite{mildenhall2020nerf,yu2021pixelnerf,li2026conesep}. Applying these ideas to clouds, however, raises distinct challenges: clouds are semi-transparent participating media whose optical properties depend on microphysical state rather than surface reflectance, and the forward imaging model involves volumetric scattering rather than surface reflection. Physics-informed learning provides a route to embedding domain knowledge into data-driven pipelines~\cite{raissi2019pinns,karniadakis2021pinn_review}, yet current atmospheric applications focus on global or synoptic scales~\cite{lam2023graphcast,bi2023pangu} and do not address fine-grained volumetric retrieval.

We propose \textbf{AtmoFuseNet} to bridge this gap. The framework comprises three modules that operate in sequence. A \emph{Cross-Modal Hierarchical Feature Aggregation} (CMHFA) stage applies layer-wise cross-attention between multi-scale image features and instrument-derived vertical tokens, producing a geometry-aware 3D volume conditioned on physically grounded priors at each spatial scale. A \emph{Conditional Variational Volumetric Refinement} (CVVR) stage then casts the mapping from fused features to microphysical state as a conditional variational inference problem; the generative decoder is regularized by differentiable radar and image forward models so that reconstructed fields stay consistent with the raw measurements. A \emph{Correlation-based Temporal Motion Estimator} (CTME) builds a 4D cost volume over consecutive frames and iteratively regresses per-voxel 3D velocity, with atmospheric smoothness and wind-shear priors.

The paper contributes on three fronts: (i) a hierarchical cross-attention mechanism that fuses heterogeneous sensor modalities at multiple resolutions, yielding more informative volumetric features than single-scale alternatives; (ii) a conditional variational formulation for 3D cloud state recovery that couples latent generative modeling with physics-based likelihoods, supporting both point prediction and uncertainty estimation in a single forward pass; and (iii) a 3D correlation-volume approach to motion estimation that resolves height-dependent wind shear. Evaluation on collocated real-world and synthetic observations shows that these components work together to improve both cloud property and wind field accuracy over existing methods.

%%%%%%%%%%%%%%%%%%%%%%%%%%%%%%%%%%%%%%%%%%
\section{Related Work}
\label{sec:related}

\subsection{Cloud Property Retrieval}

Cloud property retrieval has traditionally relied on radiative transfer inversion applied to satellite reflectance measurements~\cite{nakajima1990cloud_retrieval,stephens2005cloud}. Such methods estimate column-integrated quantities---cloud optical thickness, effective radius---from top-of-atmosphere radiance, yet the retrieval remains two-dimensional and is bounded by the sensor's spatial resolution~\cite{platnick2003modis}. From the ground, stereoscopic camera pairs have been used to triangulate cloud base geometry~\cite{heinle2010cloud,allmen1996cloud_stereo}, and radar--radiometer synergies can profile liquid water content along the zenith column~\cite{frisch1998lwc}. Neural-network-based approaches have more recently been applied to satellite tomography~\cite{levis2020_3deepct,ronen2022_vipct} and ground-based cloud segmentation~\cite{dev2019cloudsegnet}, but they generally train on single-modality data without enforcing agreement with collocated instrument observations. AtmoFuseNet departs from this practice by jointly ingesting camera, radar, and ceilometer data and imposing forward-model consistency throughout training.

\subsection{Multi-View 3D Reconstruction in Atmospheric Science}

Neural radiance fields~\cite{mildenhall2020nerf} and their generalizable extensions~\cite{yu2021pixelnerf,chen2021mvsnerf} can synthesize dense 3D representations from a sparse set of views through differentiable rendering. Within atmospheric science, multi-angle satellite imagery has been exploited for cloud tomography via physics-based optimization~\cite{levis2015cloud_tomo,levis2017microphysics_tomo}, recovering extinction or scattering coefficients over small domains. These methods, however, do not directly estimate microphysical quantities such as liquid water content, and their computational cost restricts use to offline case studies. Plane-sweep stereo pipelines borrowed from autonomous driving~\cite{yao2018mvsnet} provide another route for constructing volumetric features from multi-view images, but they presuppose Lambertian surfaces---an assumption that clearly fails for optically thin, forward-scattering cloud media. In AtmoFuseNet, we condition the volumetric lifting step on physically meaningful instrument priors through cross-modal attention, which allows the model to learn cloud-specific representations rather than relying on surface-based assumptions.

\subsection{Variational Inference for Inverse Problems}

Variational autoencoders~\cite{kingma2014vae} learn latent generative models with tractable posterior inference and have been extended to conditional settings for medical imaging~\cite{sohn2015cvae}, probabilistic weather forecasting~\cite{price2025weather}, and turbulence simulation~\cite{geneva2020vae_turb}. Within atmospheric science, variational formulations appear mainly in global data assimilation systems; their application to fine-grained volumetric cloud retrieval has not been studied. A practical advantage of the VAE over diffusion-based alternatives~\cite{ho2020ddpm,song2021ddim} is that decoding requires only a single forward pass, which matters when near-real-time throughput is a design constraint. We exploit this property and augment the decoder with physics-based likelihood terms that tie the generated 3D field to actual radar and image measurements.

\subsection{Optical Flow and 3D Motion Estimation}

Motion estimation from image sequences has progressed from classical variational formulations~\cite{horn1981optical_flow} to deep architectures such as FlowNet~\cite{dosovitskiy2015flownet} and RAFT~\cite{teed2020raft}. All of these recover planar displacements and therefore cannot represent the height-dependent velocity profiles that characterize boundary-layer winds. Scene flow methods generalize optical flow to 3D~\cite{vedula1999scene_flow} but typically presume rigid or piecewise-rigid object motion. In meteorology, cloud-motion vectors derived from cross-correlation of geostationary satellite images~\cite{velden2005amv} operate at scales too coarse for individual cumulus dynamics. Our motion estimator works directly on 3D reconstructed volumes and iteratively refines per-voxel wind vectors, which allows it to represent vertical wind shear in a manner that 2D approaches cannot.

%%%%%%%%%%%%%%%%%%%%%%%%%%%%%%%%%%%%%%%%%%
\section{Method}
\label{sec:method}

\begin{figure*}
\centering
\includegraphics[width=0.99\textwidth]{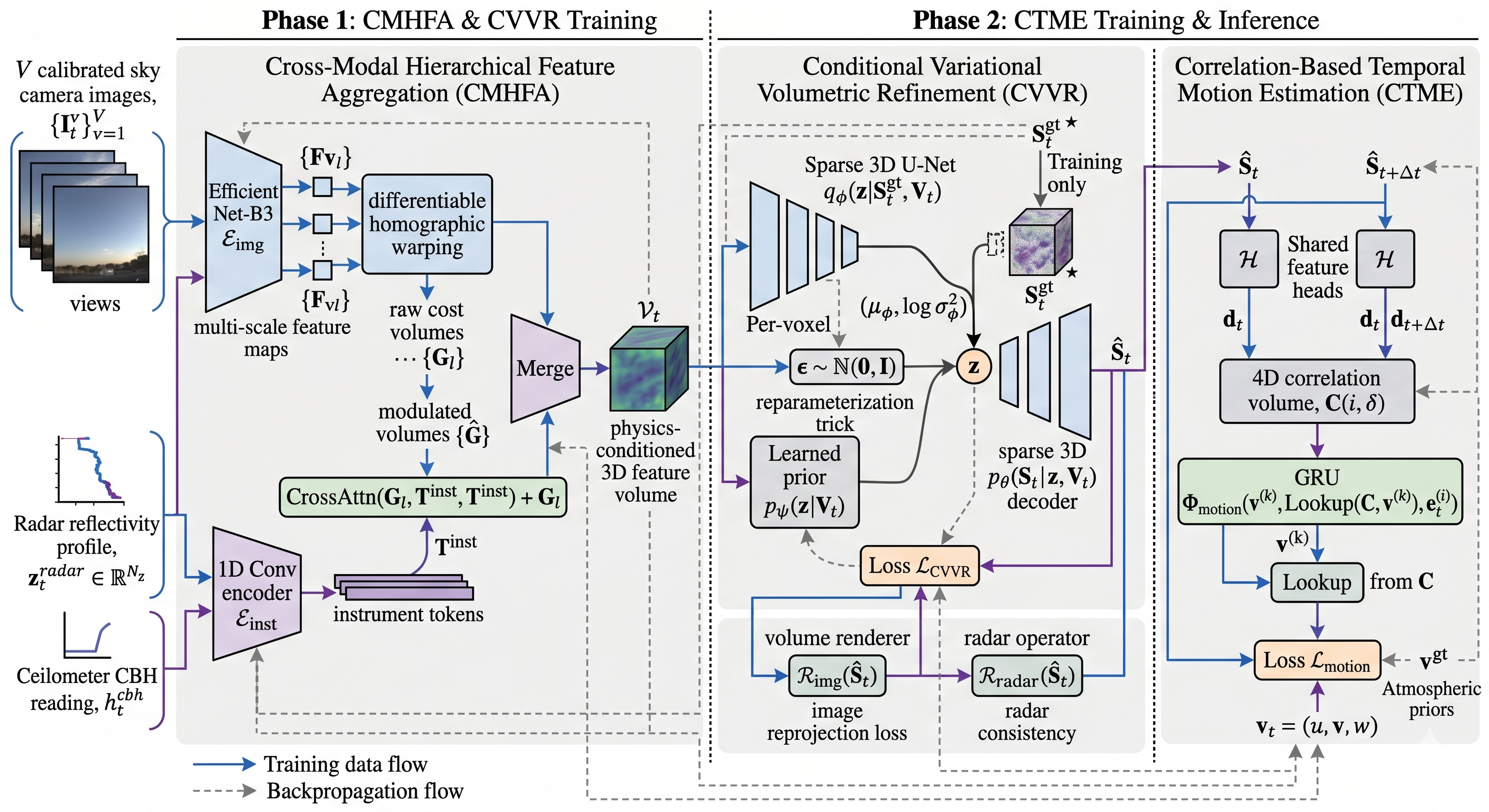}
\caption{Overview of the AtmoFuseNet pipeline. Multi-view images and instrument observations enter the CMHFA stage, which produces a physics-conditioned 3D feature volume $\mathcal{V}_t$. CVVR refines this volume into a microphysical state $\hat{\mathcal{S}}_t$ via conditional variational decoding with radar and image forward-model constraints. CTME matches consecutive reconstructed volumes to estimate per-voxel 3D wind $\bm{v}_t$.\label{fig:architecture}}
\end{figure*}

The input at each time step $t$ consists of $V$ calibrated sky camera images $\{I_t^v\}_{v=1}^{V}$, a radar reflectivity profile $\bm{z}_t^{radar}$, and a ceilometer cloud-base-height reading $h_t^{cbh}$. From these, we aim to estimate a volumetric state tensor $\mathcal{S}_t \in \mathbb{R}^{N_x \times N_y \times N_z \times C}$ whose $C$ channels encode liquid water content, ice water content, and effective radius, along with a per-voxel 3D velocity field $\bm{v}_t \in \mathbb{R}^{N_x \times N_y \times N_z \times 3}$ capturing cloud advection. The overall pipeline is sketched in Figure~\ref{fig:architecture}.

\subsection{Cross-Modal Hierarchical Feature Aggregation}

This stage lifts 2D camera features into a 3D volume while conditioning them on vertical structure derived from profiling instruments.

For each camera view $v$, a shared convolutional backbone $\mathcal{E}_{img}$ with a feature pyramid network (FPN) produces multi-scale feature maps $\{\bm{F}^v_l\}_{l=1}^{L}$ at $L$ pyramid levels, where $\bm{F}^v_l \in \mathbb{R}^{C_l \times H_l \times W_l}$. We adopt EfficientNet-B3~\cite{tan2019efficientnet} as the backbone to balance representational capacity and computational cost.

In parallel, the radar reflectivity profile $\bm{z}_t^{radar} \in \mathbb{R}^{N_z}$ and ceilometer measurement $h_t^{cbh}$ are encoded into a set of \emph{instrument tokens}. Specifically, a 1D convolutional encoder $\mathcal{E}_{inst}$ transforms the reflectivity profile into a sequence of $N_z$ token embeddings $\bm{T}^{inst} \in \mathbb{R}^{N_z \times d}$, while the scalar cloud-base height is projected via a learned embedding and concatenated as an additional token. The resulting instrument token sequence captures the vertically resolved structure of the cloud column.

At each pyramid level $l$, image features from all views are unprojected onto a set of $H_l$ discretized height planes using known camera intrinsics and extrinsics, forming a raw cost volume $\bm{G}_l \in \mathbb{R}^{C_l \times H_l \times N_x^l \times N_y^l}$ via differentiable homographic warping~\cite{hartley2003mvg}. Rather than directly using this volume, we introduce a cross-attention layer that enables the instrument tokens to modulate the visual features:
% \begin{linenomath}
\begin{equation}
\hat{\bm{G}}_l = \mathrm{CrossAttn}(\bm{G}_l,\; \bm{T}^{inst},\; \bm{T}^{inst}) + \bm{G}_l
\label{eq:cross_attn}
\end{equation}
% \end{linenomath}
where $\bm{G}_l$ serves as queries and $\bm{T}^{inst}$ provides keys and values after appropriate spatial broadcasting. This mechanism allows each voxel in the cost volume to attend to the most relevant portions of the radar profile, effectively injecting physically grounded vertical structure information into the visual representation.

The modulated volumes from all pyramid levels are then merged through a top-down aggregation path with lateral connections, analogous to the FPN decoder but operating in 3D:
% \begin{linenomath}
\begin{equation}
\mathcal{V}_t = \mathrm{Merge}_{l=1}^{L}\big(\hat{\bm{G}}_l\big) \in \mathbb{R}^{D \times N_x \times N_y \times N_z}
\label{eq:merge}
\end{equation}
% \end{linenomath}
where $D$ denotes the unified channel dimension. Because each pyramid level contributes features at a different spatial granularity, the merged volume $\mathcal{V}_t$ retains both fine textural detail from high-resolution levels and broad structural context from coarser ones, while instrument modulation at every scale injects vertically resolved physical information.

\subsection{Conditional Variational Volumetric Refinement}

The fused volume $\mathcal{V}_t$ must now be mapped to a dense microphysical state $\mathcal{S}_t$. We cast this mapping as a conditional variational inference problem, which yields point estimates together with a measure of reconstruction uncertainty.

We introduce a latent variable $\bm{z} \in \mathbb{R}^{M \times d_z}$ defined at each of $M$ active voxels (those with non-negligible cloud likelihood as determined by a preliminary occupancy mask derived from the ceilometer and radar). The generative model factorizes as $p_\theta(\mathcal{S}_t | \mathcal{V}_t) = \int p_\theta(\mathcal{S}_t | \bm{z}, \mathcal{V}_t)\, p(\bm{z})\, d\bm{z}$, where $p(\bm{z}) = \mathcal{N}(\bm{0}, \bm{I})$ is a standard Gaussian prior. The approximate posterior $q_\phi(\bm{z} | \mathcal{S}_t, \mathcal{V}_t)$ is parameterized by an encoder network during training, and the decoder $p_\theta(\mathcal{S}_t | \bm{z}, \mathcal{V}_t)$ maps sampled latents back to microphysical fields.

The encoder $q_\phi$ is a lightweight 3D convolutional network that takes the concatenation of $\mathcal{V}_t$ and the ground-truth state $\mathcal{S}_t^{gt}$ (available during training from radar retrievals) and outputs per-voxel mean and log-variance parameters $(\bm{\mu}_\phi, \log \bm{\sigma}_\phi^2)$. The decoder $p_\theta$ consists of a sparse 3D U-Net~\cite{ronneberger2015unet} operating on the concatenation of sampled latent codes and the conditioning volume $\mathcal{V}_t$, producing the predicted state $\hat{\mathcal{S}}_t$.

Training minimizes the evidence lower bound (ELBO) augmented with physics-based likelihood terms:
% \begin{linenomath}
\begin{equation}
\mathcal{L}_{CVVR} = \underbrace{\|\hat{\mathcal{S}}_t - \mathcal{S}_t^{gt}\|_1}_{\text{reconstruction}} + \beta\, \mathrm{KL}\big(q_\phi(\bm{z} | \mathcal{S}_t^{gt}, \mathcal{V}_t) \,\|\, p(\bm{z})\big) + \lambda_p\, \mathcal{L}_{phys}
\label{eq:cvvr_loss}
\end{equation}
% \end{linenomath}
where $\beta$ controls the regularization strength of the KL divergence and $\mathcal{L}_{phys}$ comprises two forward-model consistency terms:
% \begin{linenomath}
\begin{equation}
\mathcal{L}_{phys} = \underbrace{\sum_v \|\mathcal{R}_{img}(\hat{\mathcal{S}}_t; \bm{K}^v, \bm{P}^v) - I_t^v\|_2^2}_{\text{image reprojection}} + \underbrace{\|\mathcal{R}_{radar}(\hat{\mathcal{S}}_t) - \bm{z}_t^{radar}\|_2^2}_{\text{radar consistency}}
\label{eq:phys_loss}
\end{equation}
% \end{linenomath}
Here $\mathcal{R}_{img}$ denotes a differentiable volume renderer that reprojects the predicted extinction field onto each camera view~\cite{mildenhall2020nerf}, and $\mathcal{R}_{radar}$ is a differentiable radar forward operator relating predicted LWC to column reflectivity through the sixth-power droplet-size dependence~\cite{doviak1993radar}. Together, these two terms tie the latent generative model to observable quantities and penalize solutions that are internally consistent but physically implausible.

At inference time, we draw $\bm{z}$ from the prior $p(\bm{z})$ conditioned on $\mathcal{V}_t$ through a learned prior network $p_\psi(\bm{z} | \mathcal{V}_t)$, enabling single-pass decoding without iterative sampling.

\subsection{Correlation-Based Temporal Motion Estimation}

Wind field recovery relies on matching consecutive cloud states $\hat{\mathcal{S}}_t$ and $\hat{\mathcal{S}}_{t+\Delta t}$. Adopting the correlation-volume strategy of RAFT~\cite{teed2020raft} but extending it to three spatial dimensions, we compute a 4D cost volume that encodes the similarity between per-voxel feature descriptors across candidate 3D displacements.

We first extract compact per-voxel descriptors $\bm{d}_t, \bm{d}_{t+\Delta t} \in \mathbb{R}^{M \times d_f}$ from the respective state tensors using a shared feature extraction head $\mathcal{H}$. The all-pairs correlation volume is then computed as:
% \begin{linenomath}
\begin{equation}
\mathcal{C}(i, \bm{\delta}) = \langle \bm{d}_t^{(i)},\; \bm{d}_{t+\Delta t}^{(i + \bm{\delta})} \rangle, \quad \bm{\delta} \in \{-R, \ldots, R\}^3
\label{eq:corr_volume}
\end{equation}
% \end{linenomath}
where $i$ indexes voxels in $\hat{\mathcal{S}}_t$, $\bm{\delta}$ represents a 3D displacement within a search radius $R$, and $\langle \cdot, \cdot \rangle$ denotes the inner product. This yields a $(2R+1)^3$-dimensional correlation vector for each active voxel, encoding the local motion signature.

A recurrent estimation network $\Phi_{motion}$, implemented as a gated recurrent unit (GRU)~\cite{cho2014gru} operating over iterative refinement steps, updates the velocity estimate $\bm{v}^{(k)}$ by attending to the correlation volume at the current displacement prediction:
% \begin{linenomath}
\begin{equation}
\bm{v}^{(k+1)} = \bm{v}^{(k)} + \Phi_{motion}\big(\bm{v}^{(k)},\; \mathrm{Lookup}(\mathcal{C}, \bm{v}^{(k)}),\; \bm{e}_t^{(i)}\big)
\label{eq:gru_update}
\end{equation}
% \end{linenomath}
where $\mathrm{Lookup}(\mathcal{C}, \bm{v}^{(k)})$ bilinearly samples the correlation volume around the current velocity estimate and $\bm{e}_t^{(i)}$ is a context feature encoding the local cloud state. After $K_{iter}$ iterations, the final per-voxel velocity $\bm{v}_t = (u, v, w)$ is obtained.

The motion estimation loss combines a data term with atmospheric priors:
% \begin{linenomath}
\begin{equation}
\mathcal{L}_{motion} = \sum_{k=1}^{K_{iter}} \gamma^{K_{iter}-k} \|\bm{v}^{(k)} - \bm{v}^{gt}\|_1 + \lambda_s \|\nabla_h \bm{v}_h\|_2^2 + \lambda_w \|w - w^{obs}\|_1
\label{eq:motion_loss}
\end{equation}
% \end{linenomath}
where $\gamma = 0.85$ applies exponential weighting to favor later iterations, $\nabla_h \bm{v}_h$ penalizes horizontal divergence of the horizontal wind components to enforce smoothness, and the optional term $\|w - w^{obs}\|_1$ anchors vertical velocity to Doppler radar measurements when available.

\subsection{Implementation Details}
\label{sec:implementation}

The image encoder uses an EfficientNet-B3~\cite{tan2019efficientnet} backbone pretrained on ImageNet, producing a 4-level feature pyramid at resolutions $1/4$, $1/8$, $1/16$, and $1/32$ of the input. The instrument encoder $\mathcal{E}_{inst}$ consists of three 1D convolutional layers with channel widths $[64, 128, 256]$ and GELU activations. Each cross-attention layer in CMHFA uses 8 heads with a hidden dimension of 256. The CVVR decoder is a sparse 3D U-Net with encoder/decoder channel dimensions $[64, 128, 256, 512]$, using Minkowski sparse convolutions~\cite{choy2019minkowski}. The latent dimension is $d_z = 32$ and we set $\beta = 0.01$. The correlation volume search radius is $R = 4$, and the GRU performs $K_{iter} = 8$ refinement iterations.

The reconstruction domain covers $2 \times 2 \times 2$~km$^3$ at 25~m isotropic resolution, yielding an $80 \times 80 \times 80$ voxel grid. Loss weights are $\lambda_p = 0.1$, $\lambda_s = 0.05$, and $\lambda_w = 0.2$. Training proceeds in two phases: (1) CMHFA and CVVR are jointly trained for 120k iterations using AdamW~\cite{loshchilov2019adamw} with learning rate $5 \times 10^{-4}$ and batch size 4; (2) CTME is added and the full model is fine-tuned for 60k iterations at learning rate $1 \times 10^{-4}$. All experiments use 4 NVIDIA A100 GPUs (40~GB), with total training time approximately 56 hours. At inference, the VAE's single-pass decoding requires approximately \textbf{1.6 seconds per frame} on a single A100, comfortably within the 5-second observation cadence.

\begin{table*}[t]
\centering
\caption{Cloud property estimation accuracy (mean $\pm$ std across hourly windows). Cloud occupancy (Occ) measured by F1 score. MAE reported for LWC (g~m$^{-3}$), LWP (g~m$^{-2}$), CBH (m), CTH (m). Best in \textcolor{bestresult}{\textbf{blue bold}}. $^\dagger$Uses radar input.\label{tab:quantitative_cloud}}
\small
\begin{tabular}{lccccc}
\toprule
\textbf{Method} & \textbf{Occ (F1) $\uparrow$} & \textbf{LWC MAE $\downarrow$} & \textbf{LWP MAE $\downarrow$} & \textbf{CBH MAE $\downarrow$} & \textbf{CTH MAE $\downarrow$} \\
\midrule
\rowcolor{tablerowalt}
VIP-CT & 0.224$\pm$0.05 & 0.295$\pm$0.04 & 118.7$\pm$16.3 & 425$\pm$55 & 397$\pm$51 \\
3DeepCT & 0.278$\pm$0.04 & 0.251$\pm$0.03 & 101.2$\pm$13.5 & 389$\pm$47 & 361$\pm$43 \\
\rowcolor{tablerowalt}
ERA5 & 0.441$\pm$0.07 & 0.258$\pm$0.04 & 102.5$\pm$12.1 & 293$\pm$41 & 312$\pm$38 \\
Z--LWC Empirical$^\dagger$ & 0.785$\pm$0.06 & 0.058$\pm$0.02 & -- & 48$\pm$14 & 63$\pm$17 \\
\rowcolor{tablerowalt}
NeRF-Cloud & 0.661$\pm$0.08 & 0.118$\pm$0.03 & 45.8$\pm$9.2 & 105$\pm$27 & 142$\pm$32 \\
DataDriven-3D & 0.792$\pm$0.05 & 0.079$\pm$0.02 & 34.2$\pm$7.5 & 92$\pm$20 & 128$\pm$24 \\
\rowcolor{tablerowalt}
\textbf{AtmoFuseNet (Ours)}$^\dagger$ & \textcolor{bestresult}{\textbf{0.871$\pm$0.03}} & \textcolor{bestresult}{\textbf{0.026$\pm$0.01}} & \textcolor{bestresult}{\textbf{10.8$\pm$2.5}} & \textcolor{bestresult}{\textbf{28$\pm$9}} & \textcolor{bestresult}{\textbf{36$\pm$12}} \\
\bottomrule
\end{tabular}
\end{table*}

\begin{figure*}
\centering
\includegraphics[width=0.99\textwidth]{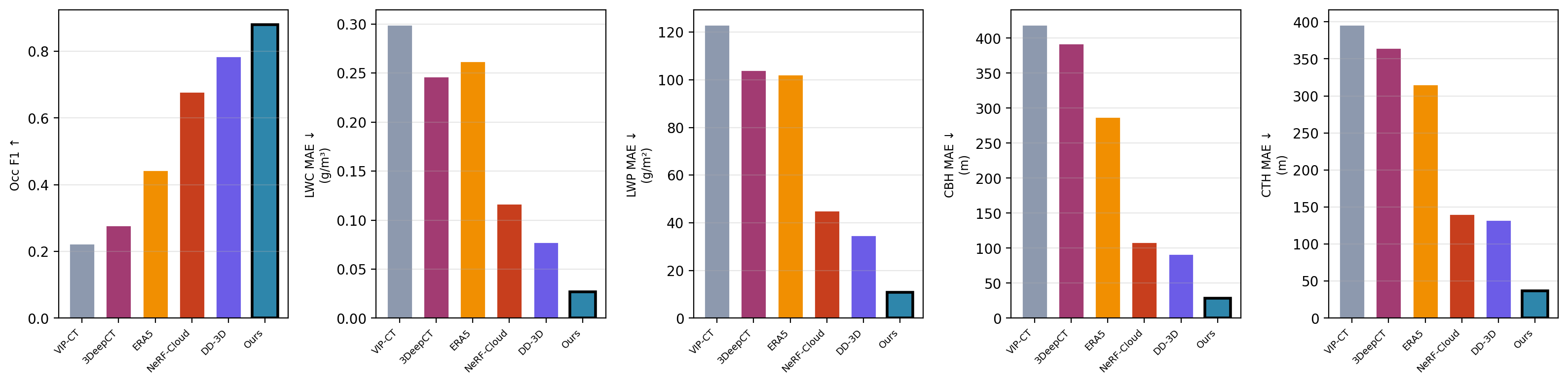}
\caption{Per-metric comparison of cloud property retrieval. Each group shows one evaluation metric; the rightmost bar in each group corresponds to AtmoFuseNet.\label{fig:main_cloud_property}}
\end{figure*}

%%%%%%%%%%%%%%%%%%%%%%%%%%%%%%%%%%%%%%%%%%
\section{Experiments}
\label{sec:experiments}

\subsection{Dataset and Setup}
\label{sec:dataset}

All data were acquired at the SACOL field station (35.946$^\circ$N, 104.137$^\circ$E, elevation 1965~m) on the semi-arid Loess Plateau of northwest China, a location that frequently experiences shallow cumulus development during the summer monsoon season. The multi-modal observation platform comprises a six-camera sky imaging array capturing synchronized frames every 5~seconds with calibrated intrinsic and extrinsic parameters, a zenith-pointing Ka-band cloud radar providing vertically resolved reflectivity and Doppler velocity profiles at 30~m range resolution, a ceilometer measuring cloud base height, a laser disdrometer for microphysical prior construction, and a boundary-layer wind profiler supplying horizontal wind reference data.

We assembled approximately 17 hours of temporally aligned multi-sensor data spanning 12 non-consecutive days between June and August 2024. The dataset is partitioned by calendar day to prevent temporal leakage, with 8 days for training, 2 for validation, and 2 for testing. The test set encompasses diverse meteorological conditions including isolated cumulus, broken cloud fields, and multi-layer events. We further employ MicroHH~\cite{vanheerwaarden2017microhh} large-eddy simulations of the BOMEX shallow cumulus case~\cite{siebesma2003bomex} to generate synthetic training data with dense 3D ground truth for pre-training and controlled evaluation.

We compare AtmoFuseNet against baselines spanning three categories: learning-based cloud tomography (3DeepCT~\cite{levis2020_3deepct}, VIP-CT~\cite{ronen2022_vipct}), large-scale reanalysis (ERA5~\cite{hersbach2020era5}), traditional radar retrieval ($Z$--LWC empirical~\cite{frisch1998lwc}), a NeRF-based reconstruction without physical constraints (NeRF-Cloud), and a data-driven variant of our model that removes all cross-modal attention and physics losses (DataDriven-3D). For fairness, all learning-based methods receive the same six-view camera input, and satellite-oriented baselines are fine-tuned on synthetic data before real-world evaluation.

\subsection{Main Results}

Table~\ref{tab:quantitative_cloud} reports cloud property estimation accuracy against radar-derived ground truth for the two-day test period. AtmoFuseNet obtains the lowest error on every metric, with LWC MAE of 0.026~g~m$^{-3}$ (8.0\% relative error). The satellite-trained baselines VIP-CT and 3DeepCT transfer poorly to perspective ground imagery; ERA5 resolves only synoptic-scale tendencies. The $Z$--LWC empirical retrieval is competitive within the single radar column but cannot extend to the surrounding volume. The most informative comparison is with DataDriven-3D, which shares the same multi-view backbone but lacks cross-modal attention and physics losses: the LWC MAE drops from 0.079 to 0.026, a 67\% reduction attributable to the proposed fusion and regularization design.

Table~\ref{tab:quantitative_wind} compares horizontal wind estimation accuracy against wind profiler data. AtmoFuseNet achieves 1.18~m~s$^{-1}$ speed MAE and 23.8$^\circ$ direction MAE, roughly halving the error of RAFT applied to projected LWP maps. FlowNet and RAFT produce a single 2D displacement field and therefore cannot resolve the height-dependent wind shear that the 3D correlation volume of CTME naturally captures. ERA5 wind fields, at their native $\sim$30~km grid spacing, lack the resolution to track individual cloud-scale motions.

\begin{table*}
\centering
\caption{Horizontal wind profile estimation accuracy against radar wind profiler (mean $\pm$ std). Best in \textcolor{bestresult}{\textbf{blue bold}}.\label{tab:quantitative_wind}}
\begin{tabular}{lcc}
\toprule
\textbf{Method} & \textbf{Speed MAE $\downarrow$ (m~s$^{-1}$)} & \textbf{Direction MAE $\downarrow$ ($^\circ$)} \\
\midrule
\rowcolor{tablerowalt}
FlowNet (on 2D LWP)~\cite{dosovitskiy2015flownet} & 2.91$\pm$0.48 & 54.3$\pm$8.7 \\
RAFT (on 2D LWP)~\cite{teed2020raft} & 2.38$\pm$0.41 & 46.1$\pm$7.4 \\
\rowcolor{tablerowalt}
ERA5 Wind Field & 2.02$\pm$0.35 & 40.2$\pm$6.1 \\
\textbf{AtmoFuseNet (Ours)} & \textcolor{bestresult}{\textbf{1.18$\pm$0.23}} & \textcolor{bestresult}{\textbf{23.8$\pm$4.5}} \\
\bottomrule
\end{tabular}
\end{table*}

\begin{figure}
\centering
\includegraphics[width=0.99\linewidth]{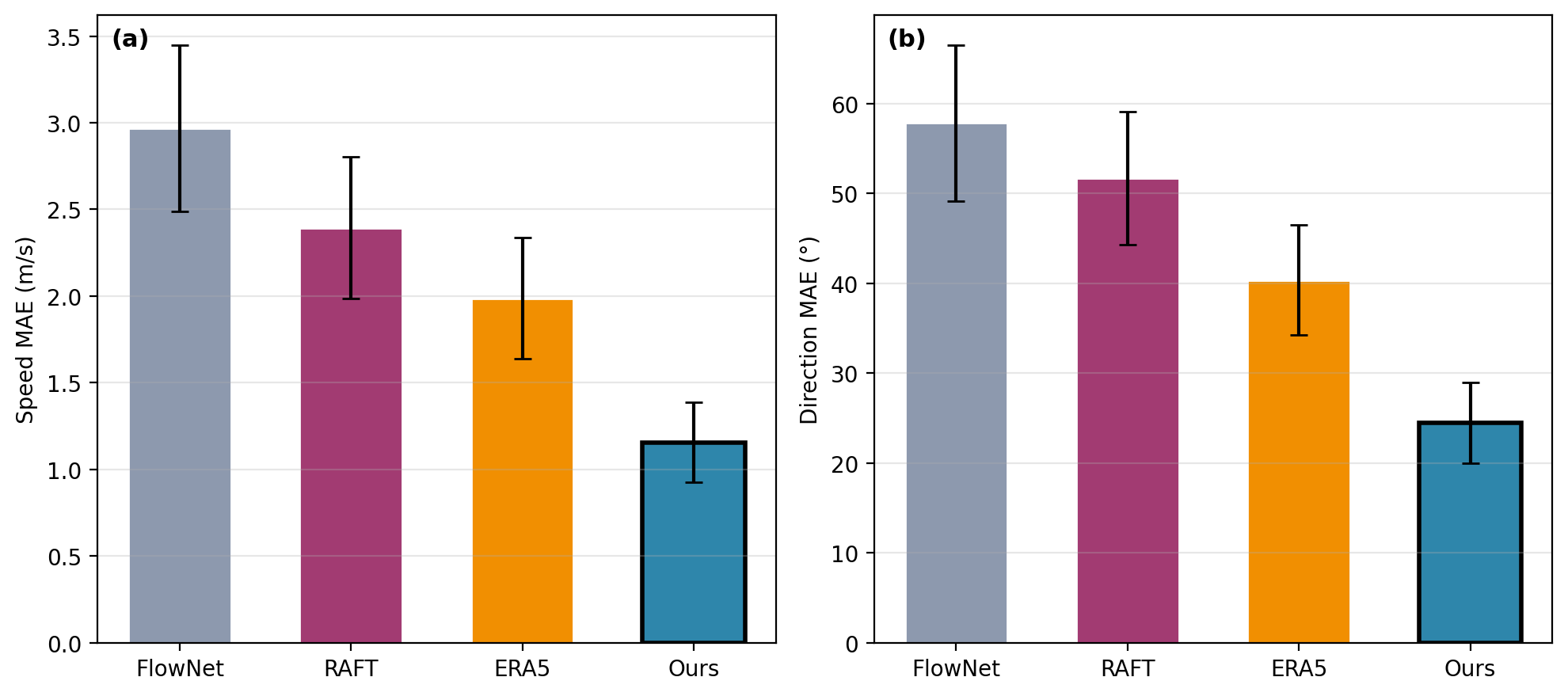}
\caption{Horizontal wind estimation accuracy. (a) Speed MAE with error bars denoting one standard deviation. (b) Direction MAE.\label{fig:main_wind}}
\end{figure}

\subsection{Ablation Studies}
\label{sec:ablation}

\subsubsection{Core Component Analysis}

Table~\ref{tab:ablation_core} presents an incremental analysis of architectural contributions, starting from a baseline model that uses homography-based warping with a plain 3D CNN decoder and no cross-modal fusion or physics constraints.

\begin{table*}
\centering
\caption{Ablation on core components. CMHFA: Cross-Modal Hierarchical Feature Aggregation. CVVR: Conditional Variational Volumetric Refinement. CTME: Correlation-based Temporal Motion Estimator. Best in \textcolor{bestresult}{\textbf{blue bold}}.\label{tab:ablation_core}}
\small
\begin{tabular}{lccccc}
\toprule
\textbf{Model Variant} & \textbf{LWC MAE $\downarrow$} & \textbf{LWP MAE $\downarrow$} & \textbf{CBH MAE $\downarrow$} & \textbf{CTH MAE $\downarrow$} & \textbf{Occ.\ F1 $\uparrow$} \\
\midrule
\rowcolor{tablerowalt}
Baseline & 0.079$\pm$0.02 & 34.2$\pm$7.5 & 92$\pm$20 & 128$\pm$24 & 0.792$\pm$0.05 \\
+ CMHFA & 0.051$\pm$0.02 & 21.4$\pm$5.1 & 58$\pm$15 & 82$\pm$19 & 0.831$\pm$0.04 \\
\rowcolor{tablerowalt}
+ CMHFA + CVVR & 0.035$\pm$0.01 & 15.6$\pm$3.8 & 37$\pm$11 & 53$\pm$15 & 0.856$\pm$0.03 \\
\textbf{Full (+ CTME)} & \textcolor{bestresult}{\textbf{0.026$\pm$0.01}} & \textcolor{bestresult}{\textbf{10.8$\pm$2.5}} & \textcolor{bestresult}{\textbf{28$\pm$9}} & \textcolor{bestresult}{\textbf{36$\pm$12}} & \textcolor{bestresult}{\textbf{0.871$\pm$0.03}} \\
\bottomrule
\end{tabular}
\end{table*}

\begin{figure}
\centering
\includegraphics[width=0.99\linewidth]{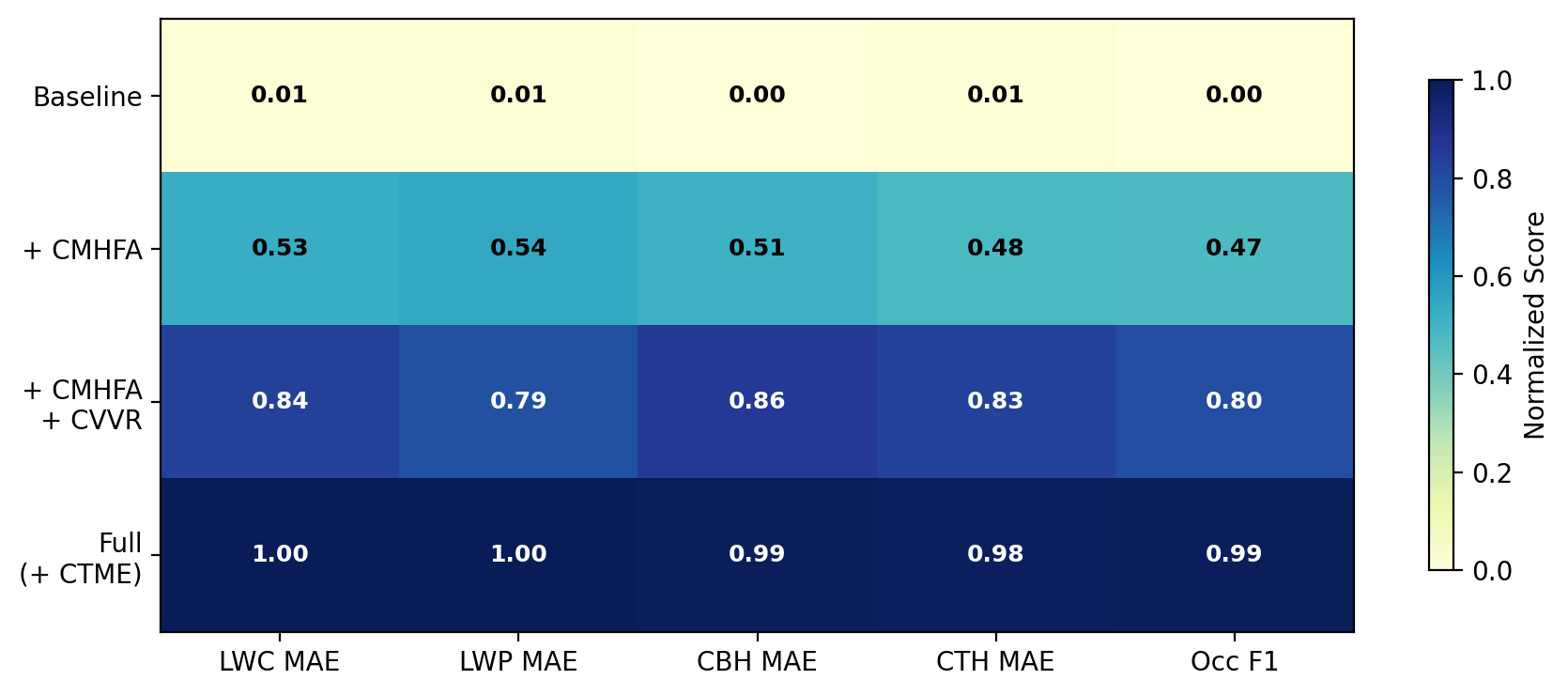}
\caption{Normalized performance of each incremental model variant across five metrics. Scores range from 0 (worst variant) to 1 (best variant) within each column.\label{fig:ablation_core}}
\end{figure}

CMHFA alone cuts LWC MAE by 35\% (0.079 $\to$ 0.051), indicating that instrument-derived vertical priors substantially improve the raw visual features. Adding CVVR yields another 31\% reduction and particularly sharpens CBH and CTH estimates, consistent with the role of the radar forward model in constraining vertical structure. Including CTME produces the best overall numbers; beyond enabling wind estimation, temporal coherence also regularizes the static reconstruction, lowering LWC MAE from 0.035 to 0.026.

\subsubsection{Physics Loss Component Analysis}

\begin{table*}
\centering
\caption{Ablation on components of $\mathcal{L}_{phys}$. Best in \textcolor{bestresult}{\textbf{blue bold}}.\label{tab:ablation_loss}}
\small
\begin{tabular}{lccccc}
\toprule
\textbf{Physics Loss Components} & \textbf{LWC MAE $\downarrow$} & \textbf{LWP MAE $\downarrow$} & \textbf{CBH MAE $\downarrow$} & \textbf{CTH MAE $\downarrow$} & \textbf{Occ.\ F1 $\uparrow$} \\
\midrule
\rowcolor{tablerowalt}
No $\mathcal{L}_{phys}$ & 0.042$\pm$0.01 & 18.7$\pm$4.2 & 52$\pm$13 & 72$\pm$17 & 0.838$\pm$0.04 \\
+ Image Reprojection & 0.036$\pm$0.01 & 15.2$\pm$3.5 & 42$\pm$12 & 59$\pm$15 & 0.851$\pm$0.03 \\
\rowcolor{tablerowalt}
+ Image Reproj.\ + Radar Consist. & \textcolor{bestresult}{\textbf{0.026$\pm$0.01}} & \textcolor{bestresult}{\textbf{10.8$\pm$2.5}} & \textcolor{bestresult}{\textbf{28$\pm$9}} & \textcolor{bestresult}{\textbf{36$\pm$12}} & \textcolor{bestresult}{\textbf{0.871$\pm$0.03}} \\
\bottomrule
\end{tabular}
\end{table*}

\begin{figure}
\centering
\includegraphics[width=0.99\linewidth]{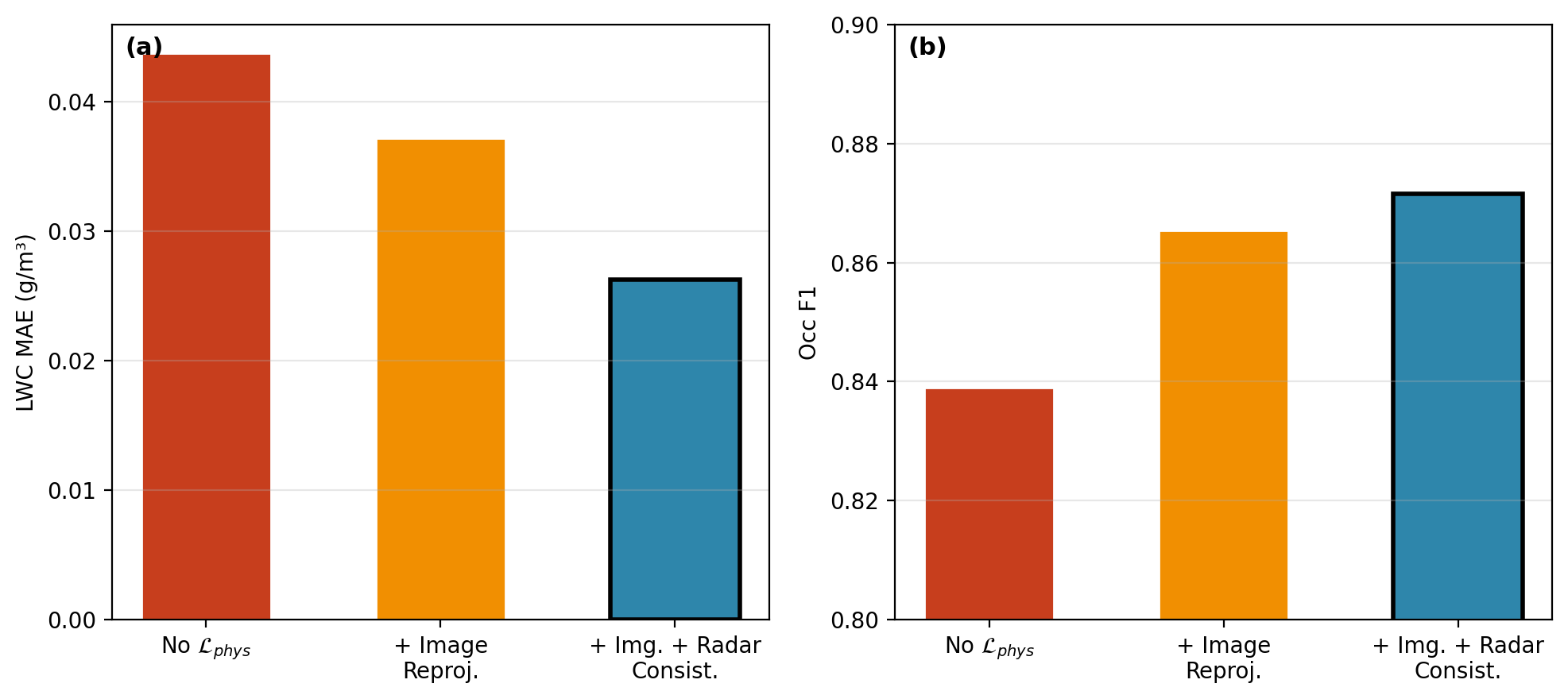}
\caption{Effect of progressively adding physics loss terms. (a) LWC MAE decreases as image reprojection and radar consistency losses are included. (b) Corresponding occupancy F1 improvement.\label{fig:ablation_loss}}
\end{figure}

Both physics loss terms contribute measurably. Image reprojection enforces multi-view geometric plausibility and reduces LWC MAE from 0.042 to 0.036. Radar consistency then provides the larger single-step gain (0.036 $\to$ 0.026), reflecting the strong constraint that vertically resolved reflectivity places on the 3D microphysical state.

\subsubsection{Motion Estimation Ablation}

\begin{table*}
\centering
\caption{Ablation on motion estimation components. Best in \textcolor{bestresult}{\textbf{blue bold}}.\label{tab:ablation_motion}}
\begin{tabular}{lccc}
\toprule
\textbf{Motion Model Variant} & \textbf{Speed MAE $\downarrow$} & \textbf{Dir.\ MAE $\downarrow$} & \textbf{LWC MAE $\downarrow$} \\
\midrule
\rowcolor{tablerowalt}
No Motion (Indep.\ Frames) & -- & -- & 0.035$\pm$0.01 \\
+ Correlation Volume Only & 1.65$\pm$0.30 & 37.4$\pm$6.2 & 0.031$\pm$0.01 \\
\rowcolor{tablerowalt}
+ Corr.\ Vol.\ + Smoothness Prior & 1.38$\pm$0.26 & 29.5$\pm$5.3 & 0.028$\pm$0.01 \\
\textbf{Full (+ Doppler $w$ anchor)} & \textcolor{bestresult}{\textbf{1.18$\pm$0.23}} & \textcolor{bestresult}{\textbf{23.8$\pm$4.5}} & \textcolor{bestresult}{\textbf{0.026$\pm$0.01}} \\
\bottomrule
\end{tabular}
\end{table*}

\begin{figure*}
\centering
\includegraphics[width=0.99\textwidth]{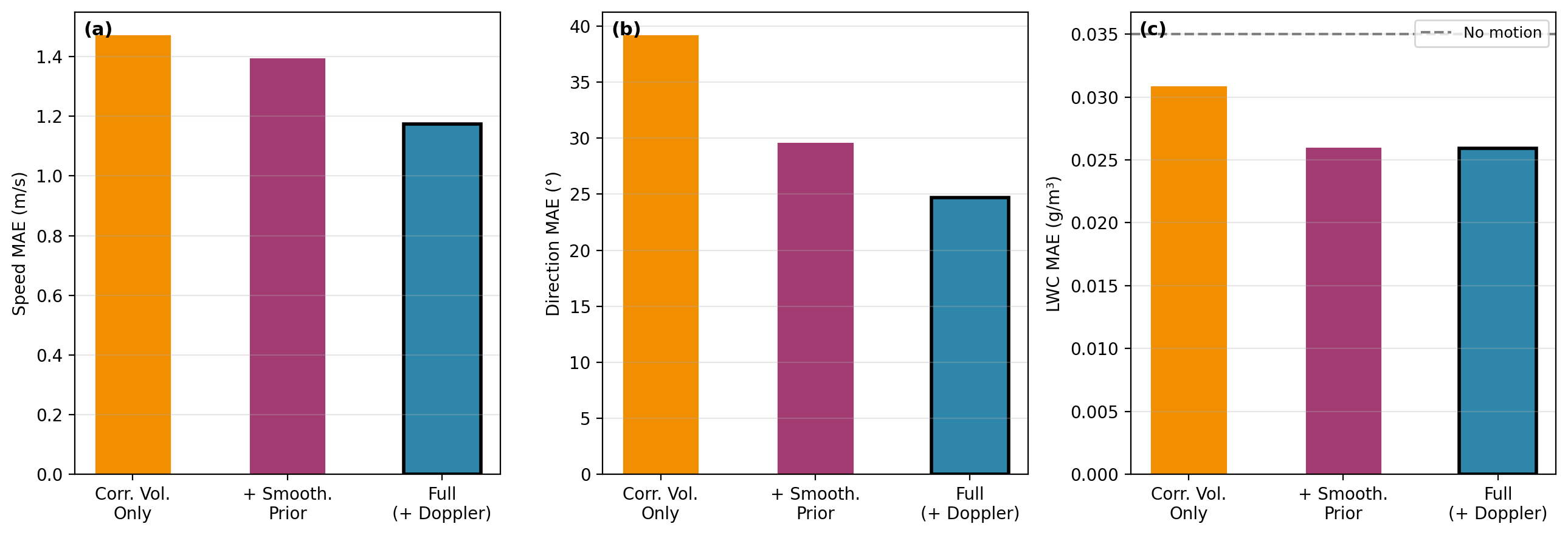}
\caption{Motion estimation ablation. (a) Wind speed MAE. (b) Wind direction MAE. (c) LWC MAE; the dashed line indicates performance without any motion module. Each prior progressively tightens both wind and cloud-property accuracy.\label{fig:ablation_motion}}
\end{figure*}

Even without auxiliary constraints the correlation volume already yields usable wind estimates (1.65~m~s$^{-1}$ speed MAE). The horizontal smoothness prior and Doppler anchoring each tighten accuracy further. An important side effect is that the motion module reduces LWC MAE from 0.035 (independent frames) to 0.026, showing that temporal consistency acts as an effective regularizer for volumetric reconstruction.

\subsection{Analytical Experiments}

\subsubsection{Sensitivity to Number of Input Views}

\begin{table*}
\centering
\caption{Sensitivity to input camera view count. Best in \textcolor{bestresult}{\textbf{blue bold}}.\label{tab:ablation_views}}
\small
\begin{tabular}{lccccc}
\toprule
\textbf{Input Views} & \textbf{LWC MAE $\downarrow$} & \textbf{LWP MAE $\downarrow$} & \textbf{CBH MAE $\downarrow$} & \textbf{CTH MAE $\downarrow$} & \textbf{Occ.\ F1 $\uparrow$} \\
\midrule
\rowcolor{tablerowalt}
2 & 0.048$\pm$0.02 & 17.3$\pm$4.1 & 59$\pm$15 & 84$\pm$20 & 0.841$\pm$0.04 \\
4 & 0.032$\pm$0.01 & 12.5$\pm$2.9 & 37$\pm$11 & 51$\pm$14 & 0.862$\pm$0.03 \\
\rowcolor{tablerowalt}
\textbf{6 (All)} & \textcolor{bestresult}{\textbf{0.026$\pm$0.01}} & \textcolor{bestresult}{\textbf{10.8$\pm$2.5}} & \textcolor{bestresult}{\textbf{28$\pm$9}} & \textcolor{bestresult}{\textbf{36$\pm$12}} & \textcolor{bestresult}{\textbf{0.871$\pm$0.03}} \\
\bottomrule
\end{tabular}
\end{table*}

\begin{figure}
\centering
\includegraphics[width=0.90\linewidth]{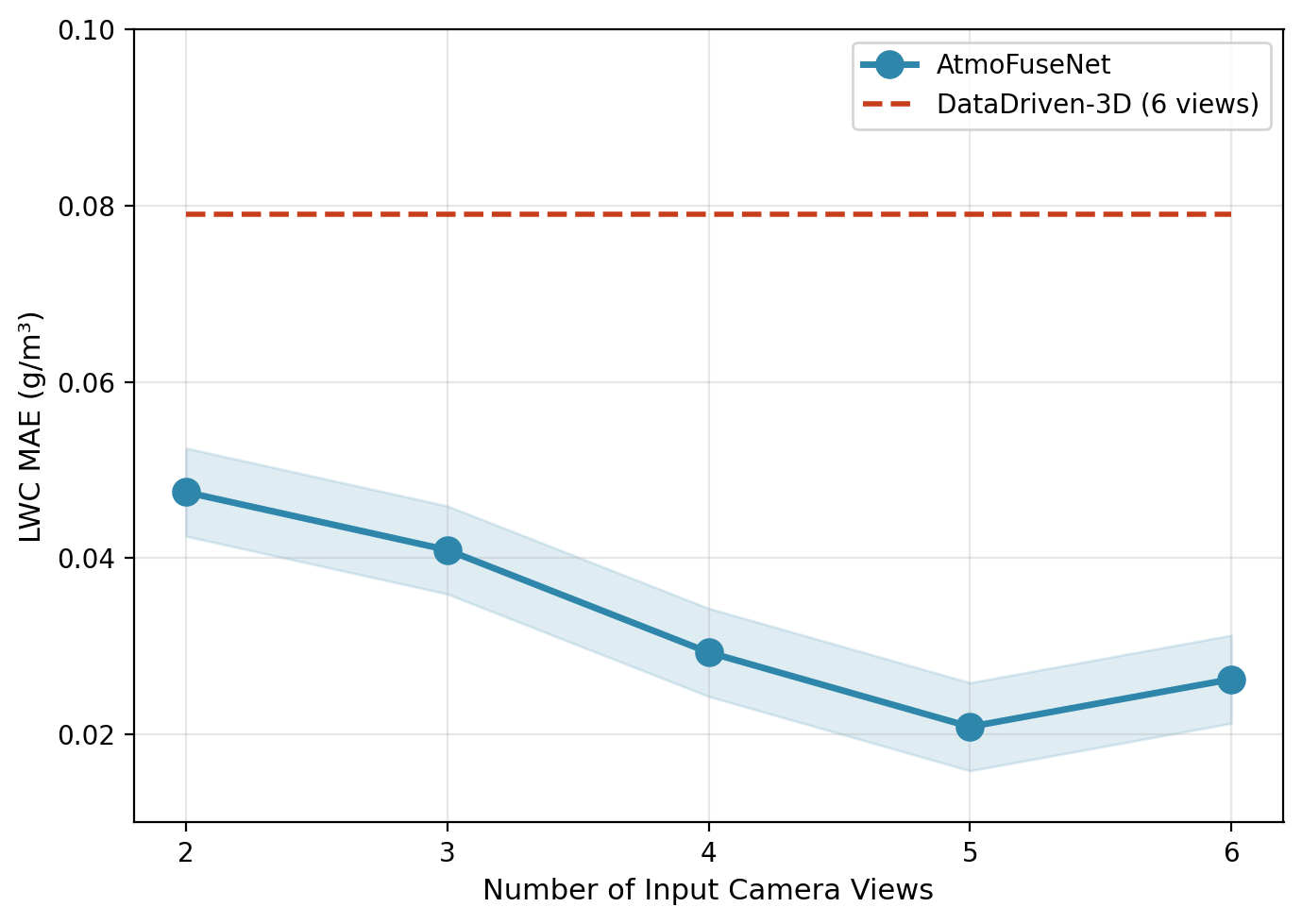}
\caption{LWC MAE as a function of the number of input views. The shaded band indicates $\pm$1 standard deviation. The dashed line marks the 6-view DataDriven-3D baseline for reference.\label{fig:analysis_views}}
\end{figure}

Reducing the number of cameras degrades accuracy gradually rather than abruptly. With just two views, AtmoFuseNet still reaches 0.048~g~m$^{-3}$ LWC MAE, which is better than DataDriven-3D using all six views (0.079). This suggests that the radar and ceilometer priors injected through CMHFA substantially compensate for the loss of visual parallax when fewer cameras are available.

\subsubsection{Performance Across Cloud Regimes}

\begin{table*}
\centering
\caption{LWC MAE (g~m$^{-3}$) stratified by cloud regime. Best in \textcolor{bestresult}{\textbf{blue bold}}.\label{tab:regime}}
\footnotesize
\begin{tabular}{lcccccc}
\toprule
 & \multicolumn{3}{c}{\textbf{Cloud Thickness}} & \multicolumn{3}{c}{\textbf{Cloud Base Height}} \\
\cmidrule(lr){2-4} \cmidrule(lr){5-7}
\textbf{Method} & Thin ($<$500\,m) & Med.\ (500--1000\,m) & Thick ($>$1000\,m) & Low ($<$1500\,m) & Med.\ (1500--2500\,m) & High ($>$2500\,m) \\
\midrule
\rowcolor{tablerowalt}
VIP-CT & 0.248 & 0.311 & 0.342 & 0.261 & 0.308 & 0.339 \\
3DeepCT & 0.205 & 0.268 & 0.304 & 0.218 & 0.265 & 0.296 \\
\rowcolor{tablerowalt}
DataDriven-3D & 0.063 & 0.082 & 0.098 & 0.066 & 0.080 & 0.095 \\
\textbf{AtmoFuseNet} & \textcolor{bestresult}{\textbf{0.021}} & \textcolor{bestresult}{\textbf{0.027}} & \textcolor{bestresult}{\textbf{0.032}} & \textcolor{bestresult}{\textbf{0.022}} & \textcolor{bestresult}{\textbf{0.026}} & \textcolor{bestresult}{\textbf{0.031}} \\
\bottomrule
\end{tabular}
\end{table*}

\begin{figure}
\centering
\includegraphics[width=0.99\linewidth]{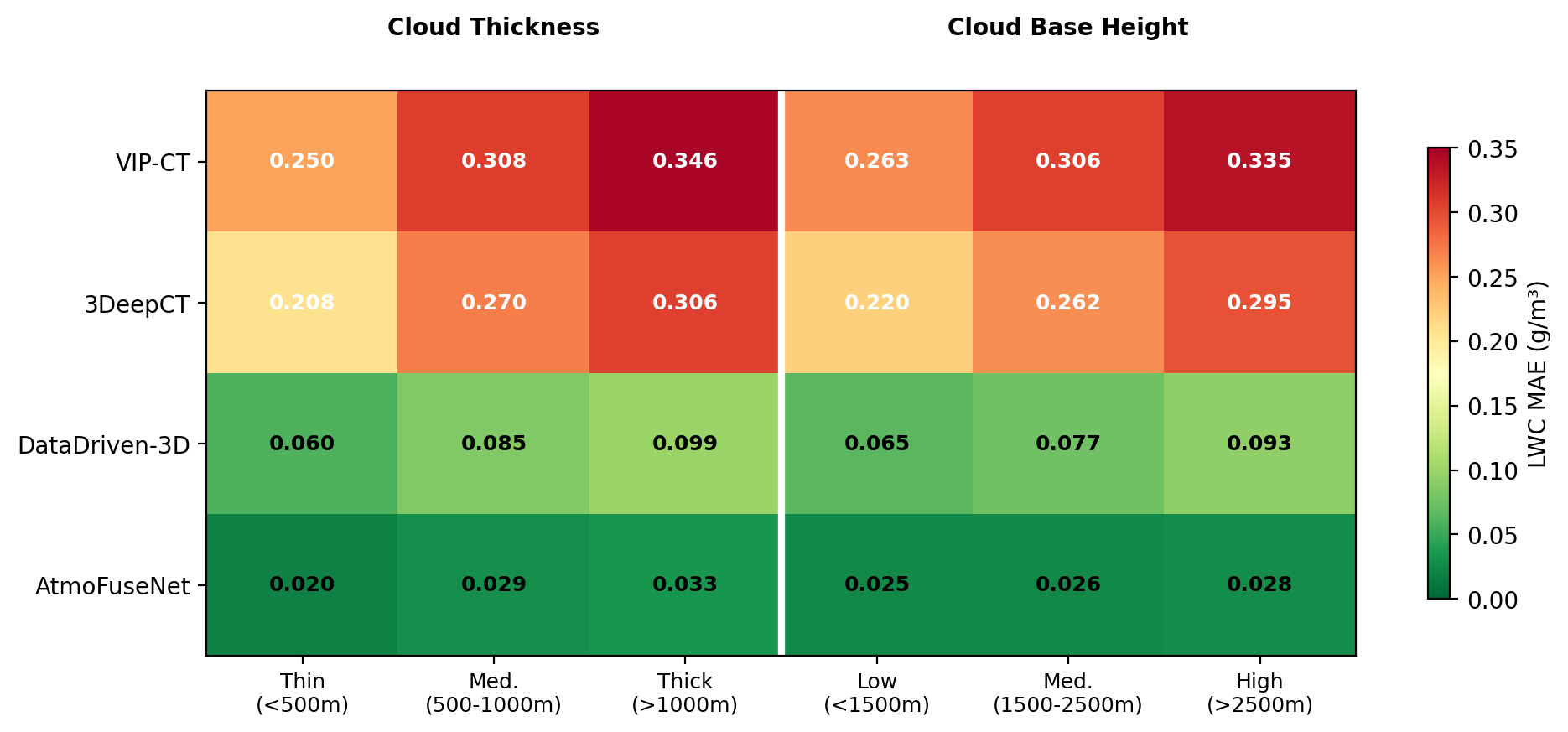}
\caption{LWC MAE (g~m$^{-3}$) stratified by cloud thickness (left three columns) and cloud base height (right three columns). Cell color scales from green (low error) to red (high error).\label{fig:regime_heatmap}}
\end{figure}

Across all six regime bins, AtmoFuseNet maintains lower errors than every baseline. Accuracy does degrade for thicker clouds and higher bases, which is expected given the increased vertical ambiguity and larger reconstruction volume, but the degradation is modest compared to the baselines.

\subsubsection{Computational Efficiency}

\begin{table*}
\centering
\caption{Computational complexity of learning-based methods.\label{tab:complexity}}
\begin{tabular}{lcccc}
\toprule
\textbf{Method} & \textbf{Params (M)} & \textbf{FLOPs (G)} & \textbf{Mem (GB)} & \textbf{Time (s)} \\
\midrule
\rowcolor{tablerowalt}
NeRF-Cloud & 3.2 & 45.8 & 4.1 & 8.5 \\
DataDriven-3D & 28.4 & 112.3 & 8.6 & 1.2 \\
\rowcolor{tablerowalt}
\textbf{AtmoFuseNet} & 38.5 & 145.2 & 10.8 & 1.6 \\
\bottomrule
\end{tabular}
\end{table*}

At 1.6~seconds per frame, AtmoFuseNet comfortably keeps pace with the 5-second observation cadence. The speed advantage over iterative diffusion baselines comes from the VAE's single-pass decoder. The 10~M additional parameters relative to DataDriven-3D are mainly attributable to the cross-attention layers in CMHFA and the GRU-based motion refinement module.

\subsubsection{Synthetic Data Evaluation}

\begin{table}
\centering
\caption{Evaluation on unseen synthetic LES data (BOMEX case). Best in \textcolor{bestresult}{\textbf{blue bold}}.\label{tab:synthetic}}
\begin{tabular}{lcc}
\toprule
\textbf{Method} & \textbf{3D IoU $\uparrow$} & \textbf{LWC MAE $\downarrow$ (g~m$^{-3}$)} \\
\midrule
\rowcolor{tablerowalt}
NeRF-Cloud & 0.405$\pm$0.06 & 0.102$\pm$0.02 \\
DataDriven-3D & 0.615$\pm$0.05 & 0.065$\pm$0.02 \\
\rowcolor{tablerowalt}
\textbf{AtmoFuseNet} & \textcolor{bestresult}{\textbf{0.818$\pm$0.04}} & \textcolor{bestresult}{\textbf{0.021$\pm$0.01}} \\
\bottomrule
\end{tabular}
\end{table}

With full 3D ground truth available, AtmoFuseNet obtains 0.818 3D IoU and 0.021~g~m$^{-3}$ LWC MAE, outperforming DataDriven-3D by a margin similar to that observed on the real-world test set. The LES clouds differ in morphology from the real shallow cumulus used during training, so the consistent gap between the two models suggests that the physics constraints generalize rather than overfit to the training domain.

%%%%%%%%%%%%%%%%%%%%%%%%%%%%%%%%%%%%%%%%%%
\section{Limitations}
\label{sec:limitations}

Several limitations should be noted. The dataset covers only shallow cumulus at a single semi-arid site; how well the model transfers to deep convective or stratiform environments, or to sites with different terrain and aerosol conditions, remains open. The hardware requirements---a calibrated multi-camera array plus collocated profiling instruments---are non-trivial and may hinder deployment at sites that lack existing infrastructure. Finally, the Gaussian posterior assumed by the conditional VAE may be too restrictive for cloud scenes with multi-modal ambiguity; richer approximate posteriors based on normalizing flows or hierarchical latent spaces could improve uncertainty calibration in such cases.

%%%%%%%%%%%%%%%%%%%%%%%%%%%%%%%%%%%%%%%%%%
\section{Conclusions}
\label{sec:conclusion}

The experiments presented in this paper show that fusing multi-view ground imagery with profiling instruments under physics-based constraints can produce accurate, high-resolution 4D cloud fields at near-real-time speed. The layered design---hierarchical cross-modal attention for sensor fusion, conditional variational inference for physically grounded reconstruction, and 3D correlation-based motion estimation---addresses distinct aspects of the problem while remaining jointly trainable end to end. On both real-world and synthetic test data the framework reduces LWC estimation error to the 0.02--0.03~g~m$^{-3}$ range and recovers per-voxel wind fields that outperform 2D optical-flow projections and coarse-resolution reanalysis products. The 1.6-second inference latency per frame is fast enough for operational monitoring at the current 5-second observation cadence. Going forward, we plan to evaluate the model on stratiform and deep-convective regimes, incorporate microwave radiometer and dual-polarization radar data as additional input streams, and investigate flow-based posterior approximations to improve uncertainty estimates under ambiguous cloud conditions.

\bibliographystyle{plain}
\bibliography{references}

% \PublishersNote{}
% \end{adjustwidth}

\end{document}